\documentclass[manuscript]{acmart}

\newcommand{\crazyLO}{
	\begin{sideways}\begin{minipage}{30pt}
	{\emph{Land\\ \emph{Only}}}
	\end{minipage}\end{sideways}}

\newcommand{\crazyLA}{
	\begin{sideways}\begin{minipage}{72pt}
	{\emph{Land \emph{AND} Air Units}}
	\end{minipage}\end{sideways}}
	
\newcommand{\crazyAO}{
	\begin{sideways}\begin{minipage}{60pt}
	{\emph{Air Units \emph{Only}}}
	\end{minipage}\end{sideways}}

\newcommand{\REDACT}[1]{$\Box\Box\Box$} 

\newcommand{\redactCollege}[1]{[a U.S. University]}  

\newcounter{boldifyCounter}
\newcounter{fixmeSectionCounter}
\newcounter{fixmeTotalCounter}
\makeatletter
\@addtoreset{fixmeSectionCounter}{section}
\@addtoreset{fixmeSectionCounter}{subsection}
\@addtoreset{boldifyCounter}{section}
\@addtoreset{boldifyCounter}{subsection}
\makeatother

\newcommand{\boldify}[1]{}
\ifdefined\boldifyON
	\renewcommand{\boldify}[1]{
        \par\noindent
		\stepcounter{boldifyCounter}
		\textbf{{\color{green}**}
		~\arabic{section}.\arabic{subsection}.\arabic{boldifyCounter}
		: #1} 
	}
\fi 

\newcommand{\reportOnFIXME}{%
    \newcount\iterCounter
    \iterCounter=1
    \newcount\endCounter
    \endCounter=\totvalue{fixmeTotalCounter}
    \advance \endCounter +1
    There are 
    {\color{red}\total{fixmeTotalCounter}} 
    FIX\_ME\\
    links:
    \loop
        \hyperlink{fixTag\the\iterCounter}{\#\the\iterCounter}
        \advance \iterCounter +1
    \ifnum \iterCounter < \endCounter
    \repeat
}

\newcommand{\FIXME}[1]{} 
\ifdefined\fixmeON
	\renewcommand{\FIXME}[1]{\par\noindent
		\stepcounter{fixmeSectionCounter}\stepcounter{fixmeTotalCounter}
		{\color{red}\fbox{\color{black}
			\parbox{.965\linewidth}{
				\textbf{\hypertarget{fixTag\thefixmeTotalCounter}{FIXME}	\arabic{section}.\arabic{subsection}.
        		\arabic{fixmeSectionCounter} (\color{red}
        		\#\arabic{fixmeTotalCounter}):} #1}}
        }
	}
\fi 

\newcommand{\FIXED}[1]{}
\ifdefined\fixmeON
	\renewcommand{\FIXED}[1]{\par\noindent%
		{\color{black}\fbox{\color{black}%
			\parbox{.99\columnwidth}{%
				\color{blue}#1}}%
        }
	}
\fi 

\newcommand{\draftStatus}[1]{}
\ifdefined\draftStatusON
	\renewcommand{\draftStatus}[1]{
        \hfill **#1
	}
\fi 

\usepackage{calc, enumitem} 
\usepackage{soul}
\usepackage{xcolor, colortbl}
\usepackage{multirow}
\usepackage{rotating}
\usepackage{subcaption}
\usepackage{stfloats}
\usepackage{wasysym} 
\usepackage{totcount}\regtotcounter{fixmeTotalCounter}
\usepackage{datetime2} 
\usepackage{float}
\usepackage{placeins}

\copyrightyear{2023} 
\acmYear{2023} 
\setcopyright{none}\acmConference[]{Scholarly Paper for M.S.}{Submitted August 7}{Penn State University}
\acmBooktitle{Scholarly Paper for M.S., submitted August 7, 2023 to Penn State University}
\acmPrice{15.00}
\acmDOI{10.1145/3490099.3511115}
\acmISBN{978-1-4503-9144-3/22/03}

\begin{document}

\title{Experiments with Encoding Structured Data for Neural Networks}

\author{Sujay Nagesh Koujalgi}
\email{snk5290@psu.edu}
\affiliation{%
  \institution{%
  Penn State University}
  \streetaddress{Westgate Building}
  \city{University Park}
  \state{PA}
  \country{USA}
  \postcode{16802}
}

\author{Jonathan Dodge}
\email{jxd6067@psu.edu}
\affiliation{%
  \institution{%
  Penn State University}
  \streetaddress{Westgate Building}
  \city{University Park}
  \state{PA}
  \country{USA}
  \postcode{16802}
}

\begin{abstract}
  The project's aim is to create an AI agent capable of selecting good actions in a game-playing domain called Battlespace.
  Sequential domains like Battlespace are important testbeds for planning problems, as such, the Department of Defense uses such domains for wargaming exercises.
  The agents we developed combine Monte Carlo Tree Search (MCTS) and Deep Q-Network (DQN) techniques in an effort to navigate the game environment, avoid obstacles, interact with adversaries, and capture the flag.
  This paper will focus on the encoding techniques we explored to present complex structured data stored in a Python class, a necessary precursor to an agent.
\end{abstract}


%

\maketitle

\section{Introduction}

\boldify{Sequential domains are important because they encompass planning problems.
One kind of planning problem that is of particular import to many people is wargaming.
What is wargaming?}

Decisionmaking in sequential domains entails considering possible future states that may arrive as a consequence of decisions in the present.
This is the essence of a planning problem, and one instance of planning problems of particular import is wargaming, which is a simulated military exercise to test strategies and operational plans in a controlled environment.
Within the context of planning problems, it is easy to envision a variety of applications for AI, ranging from decision support systems (DSS), intelligent opponents, scenario generation, and so on.
This work will focus primarily on agents usable for the first two applications: DSS and intelligent opponents.
As a result, we seek not only agents that can select a good quality action, but also agents that can analyze multi-player interactions, adapt rapidly to changing conditions, and provide interpretable insights.

\boldify{DoD introduced Battlespace as a platform for wargaming.
It includes key features like partial observability, the capacity to have multiple players controlling different units in both cooperative and competitive fashion, and so on.
Our goal is primarily to support in-situ decision support, and to a lesser extent retrospective analysis}

The U.S. Department of Defense introduced Battlespace~\cite{Hare_etal21} as a platform for wargaming~\cite{hung2022final}.
It includes key features like partial observability~\cite{russell2016artificial}, multi-player modes (both cooperative and competitive modes), multi-level terrains, playable units with different properties, malleability to test AI agents, and so on.
Our goal is to train an agent capable enough to be the engine behind a DSS intended to work primarily \emph{in-situ}, with a goal of making better decisions while wargaming, and secondarily during \emph{retrospective analysis}, with a goal of making better decisions in the next wargaming session.

\boldify{Battlespace domain has the following key challenges}

The Battlespace domain has several key challenges, due to its complex structure, which we will detail in Section~\ref{sec:overview}. 
Among the important challenges to handle are the following.
\begin{itemize}
\item Action Space Sparsity: Battlespace has a literal ``do nothing,'' but also has many other moves that do not impact the game state much (spinning around).
\item State Space Sparsity: Battlespace uses a grid structure and most of the cells are unoccupied.
\item Large Input Space: Battlespace's structured data has a lot of properties, which leads to a proliferation of layers required to encode the state class and also leads to computational bottlenecks.
\item Multi-Agent Collaboration Strategies: Battlespace includes multiple players who each control multiple units, learning strategies in this landscape is difficult, which worsens the computational bottlenecks
\end{itemize}

\boldify{and now we specify the scope of the paper and contribution}

This project involves experiments with encoding the structured data from the Battlespace domain, training the AI agent on a simulated game environment.
Our goal in training was to enable the agent to learn how to navigate the environment, avoid obstacles, interact with adversaries, and capture the flag.
By using our encoding algorithm, we can convert observations of the board, present in the form of structured data (in this case, a state class), into a tensor.
Having done so, we can present complex states as input to a neural network (NN), which acts as a function approximator to learn game dynamics.
In this project report, we will provide an overview of the project, describe the methods we developed, and present the results obtained.

\section{Overview of Battlespace}
\label{sec:overview}

\begin{figure}
    \centering
    \includegraphics[width=0.8\textwidth]{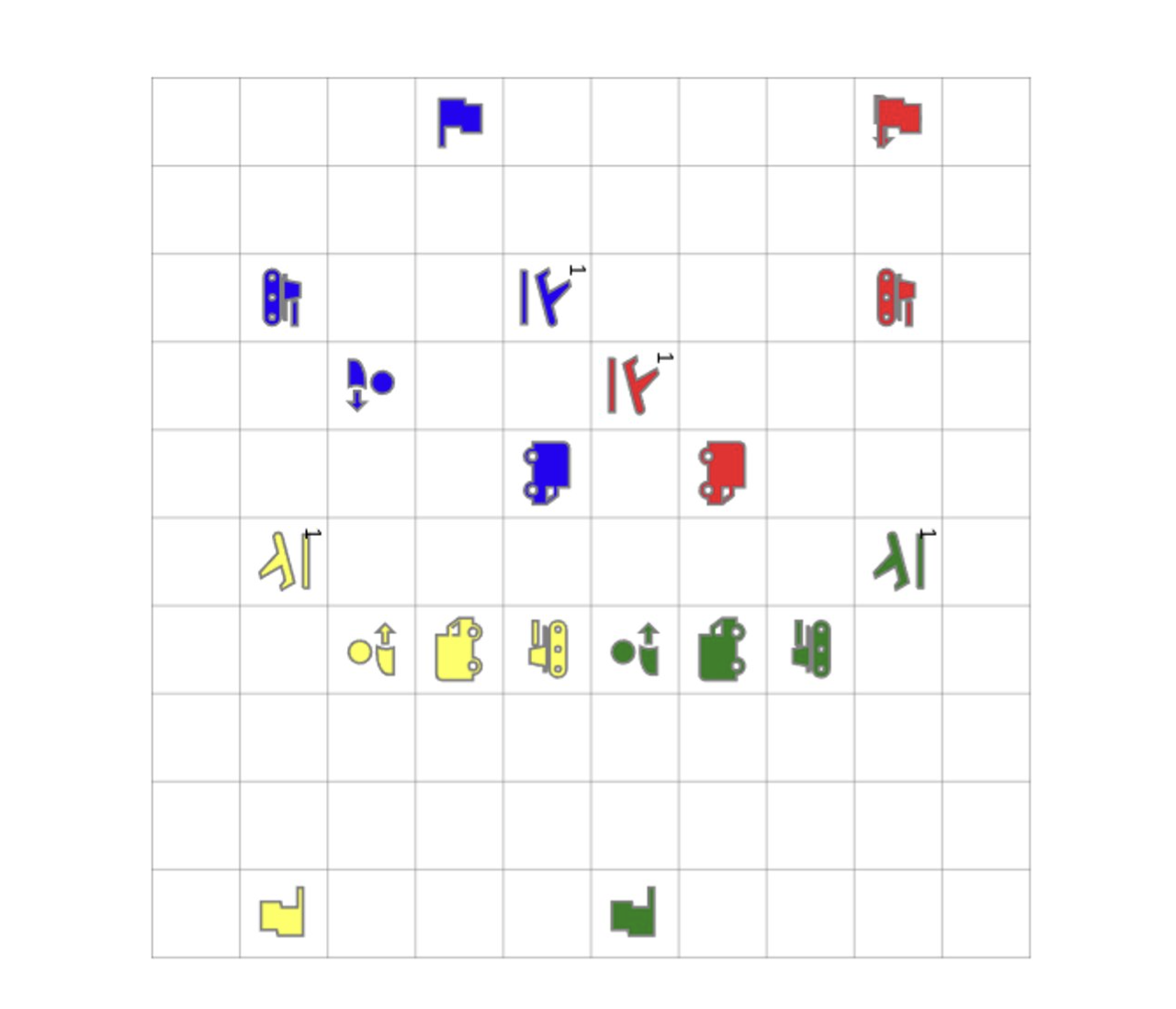}
    \caption{An example Battlespace board, snapshot from Mid Game:
    Four players represented by different colors are playing the game.
    Purple and Red units are in one team, deployed in the northern half of the board.
    Yellow and Green are in the second team, deployed on the southern half of the board.
    We can see that players have employed a safe playing strategy by placing their flags at the edge of the board as far away from the enemy territory as possible, with units in front of the flag.
    In this figure, we see both the air layer and the ground layer superimposed because that is how the built-in UI renders the game.}
    \label{fig:board}
\end{figure}

\boldify{What are modes and objectives of the game}

The Battlespace board game is a complex game with two modes, namely Annihilation and Capture the Flag.
The goal of Annihilation is to completely destroy enemy units to win the game.
Capture the Flag is an extension of Annihilation. This game is won either by destroying all enemy units or reaching the enemy's flag.
We estimate the state space of the game to be incredibly large, with around $50,000$ possible states based on the number of units and unit properties (See Appendix~\ref{appendix:ssc}). 

\boldify{what is the spatial element of the game?}

Our version of the game is playable on a board of dimensions of $10\times11\times2$, where the first two dimensions are the width and the length of the board shown in Figure~\ref{fig:board}. 
The third dimension shows if the object is on land or in the air.
We can imagine the aerial dimension as a grid of shape $10\times11$ placed above the land grid of $10\times11$.
Players only have imperfect information due to partial observability, as players cannot see the opponent's pieces until they are within the visible range of x squares in each ordinal direction.

\boldify{what are the phases of the game?}

Each team can have one or two players and must strategically place their units, including a flag, soldier, tank, truck, and airplane, on their segment of the board in the \emph{deployment phase} (detailed later in Appendix Figures~\ref{fig:deployment} and \ref{fig:deployment2}. 
Note that game parameters like board size and number of units are mutable and we have made simplifying assumptions for our experiments which is mentioned in Section~\ref{sec:experiments}.
After the deployment phase, the game begins.

\boldify{related to phases, what are the temporal aspects of this game}

We define \textbf{move} as a selection of an action for a unit, \textbf{turn} as a collection of moves for all units for a player and \textbf{round} as a collection of all the turns. 
The players enter the moves for their units on the terminal synchronously and all the moves are compiled at the end of a round. 
The simulator resolves all players' actions at the end of a round and updates the game state.
The UI updates after each round and players can see the changes as a consequence of their previous turn.

\boldify{and what are the units}

\begin{figure}
    \centering
    \includegraphics[width=\linewidth]{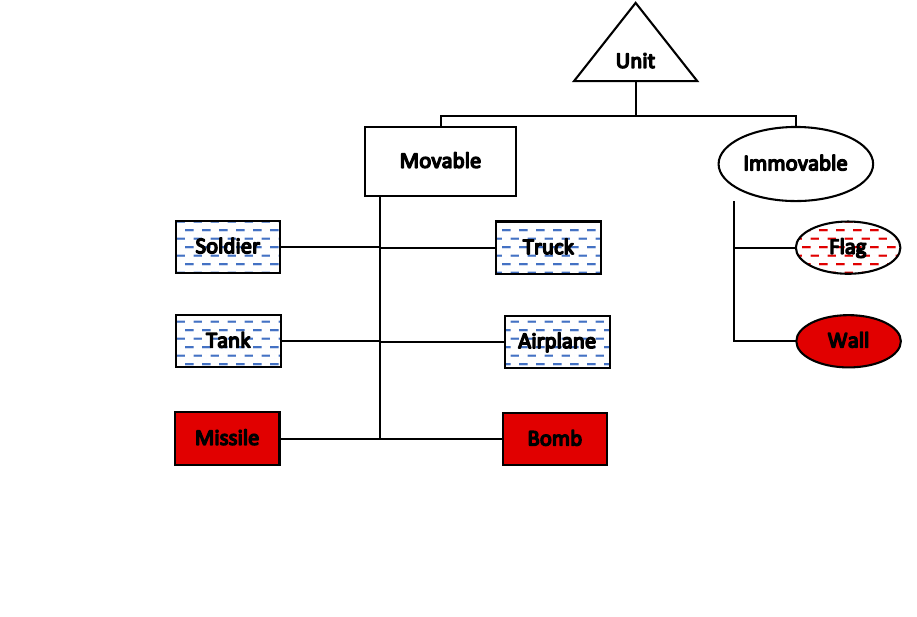}
    \caption{Class hierarchy used in Battlespace.
    We represented the \emph{abstract} classes (each having no fill color) via the following shapes: 
    Unit with a triangle, Movable with a rectangle, and Immovable with an oval.
    The \emph{player-deployed} Units are textured, meaning those with a solid fill are not deployed by the user during the deployment phase.
    The \emph{player-controlled} units have a blue fill color, meaning those with a red fill color are not under user control after the deployment phase.
    }
    \label{fig:hierarchy}
\end{figure}

The units in the game can be classified into movable and non-movable units. 
The movable units can change their position (and possibly orientation. Reference Fig 2) in the game.
Soldiers, tanks, trucks, airplanes, missiles, and bombs are the exhaustive list of movable units (missiles and bombs are not player-deployed or player-controlled). By player-deployed, we mean the units are deployed in the game deployment phase by the player. However, missiles and bombs are shot by users when they select actions..
Figure~\ref{fig:hierarchy} shows the class hierarchy.
Once immovable units are placed on the map, they cannot be moved or re-oriented. 
In the default setting, each player must place 5 units on the board that include a soldier, tank, truck, plane, and flag in their quadrant of the grid, resulting in a board that might look like the one shown in Figure~\ref{fig:board}. 
In our case, the game places 10 walls at the center of the grid on land terrain with an opening in the middle to set up a breaching scenario~\cite{hung2022final}. 
In other words, to go from one side of the board to another, there is only one possible grid unit on the land terrain which is at the center of the board. 
This contrasts with objects in the air like airplanes and missiles shot in the air that can move over the walls.

\boldify{and what can the units DO?}

\begin{table}
    \centering
    \begin{tabular}[h]{
    p{0.05\linewidth}|
    l|
    p{0.6\linewidth}} 
        & \textbf{\emph{Actions}} 
        & \textbf{\emph{Definitions}} 
        \\\hline \hline
        
        \multirow{2}{\linewidth}{\crazyLO}
        & ram 
        & Ram in the direction of current orientation.
        Enemy unit present in the location of the rammed square will be destroyed.
        If a wall is present, then the unit ramming will be destroyed. 
        \\\cline{2-3}
        
        &    advance1 
        &	Move one square forward toward the current orientation.
        \\\hline \hline
        
        \multirow{10}{*}{\crazyLA}
        & shoot
        & Shoot a missile in the direction of the current orientation.
        \\\cline{2-3}

        & doNothing
        & Take no action, maintaining position, orientiation, etc.
        \\\cline{2-3}
        
        & turn-135
        & \multirow{9}{\linewidth}{The unit takes its current orientation as the x-axis reference and then rotates by the specified values relative around the \texttt{height} axis.}
        \\\cline{2-2}
        
        &turn-90&\\\cline{2-2}
        &turn-45&\\\cline{2-2}
        &turn0&\\\cline{2-2}
        &turn45&\\\cline{2-2}
        &turn90&\\\cline{2-2}
        &turn135&\\\cline{2-2}
        &turn180&
        \\\hline \hline
        
        \multirow{14}{*}{\crazyAO}
        & bomb
        & Drop a bomb downwards.
        \\\cline{2-3}
        
        & advance0,-2
        & \multirow{13}{\linewidth}{Advance by $(a,b)$ squares, while maintaining the same orientation.}
        \\\cline{2-2}
        
        &advance-1,-1&\\\cline{2-2}
        &advance0,-1&\\\cline{2-2}
        &advance1,-1&\\\cline{2-2}
        &advance-2,0&\\\cline{2-2}
        &advance-1,0&\\\cline{2-2}
        &advance0,0&\\\cline{2-2}
        &advance1,0&\\\cline{2-2}
        &advance2,0&\\\cline{2-2}
        &advance-1,1&\\\cline{2-2}
        &advance0,1&\\\cline{2-2}
        &advance1,1&\\\cline{2-2}
        &advance0,2&
        \\\hline
        
    \end{tabular}
    \caption{List of actions available to both land and air units in Battlespace, with a description of each action.
    For each playable unit, the player can choose an action based on it being a land unit like Soldier, Tank, or Truck (top two blocks of actions) or if it's an air unit like an Airplane (bottom two blocks of actions).
    Each playable unit will perform one of these actions in one turn.}
    \label{tab:actionSpace}
\end{table}
 
Some of the movable units of the game have a different set of actions, shown in Table~\ref{tab:actionSpace}. 
Land units (soldier, tank and truck) have 12 actions: moving forward, rotating, shooting, etc. 
Air units (airplanes) have 26 actions: most of the land actions but more flexible in movement. 
``Projectile units'' (bombs, missiles) have only 1 action which is to continue moving in a direction once fired.
This action is not controlled by the user and is encoded as a property of projectile units. 
In particular, missiles continue in the direction of their projection and bombs drop downwards till they reach land. 
Walls can stop projectiles, but only if they are on the ground layer, meaning missiles shot by planes will go above the walls.
Immovable units (flags, walls) stay in their deployment location till the game is over and cannot be destroyed.

\boldify{How do we decide the rewards?}

We define the reward system for this game to be +1 for a win, 0 for a draw, and -1 for a loss.
While we used a ``sparse reward'' approach, there are multiple ways to shape the rewards.
For example, we could define rewards for different milestones like positive rewards for capturing an enemy piece, moving towards the enemy territory, or spotting the enemy flag; with negative rewards for losing a piece, being spotted in enemy territory, or having the enemy spot your flag.
Note that these kinds of shaping rewards would lend themselves naturally to rewards decomposition, as used by Anderson et al.~\cite{anderson2019explainingMereMortals, anderson2020mereMortals}.
In previous work on Battlespace~\cite{Hare_etal21, hung2022final}, the authors suggested that reward shaping biased the agent to choose to shoot repeatedly over other actions.
To avoid this and other biases that might come from ill-specified shaping reward (e.g., the famous boat race example), we stuck to the sparse rewards at the end of a full trajectory listed previously.

\boldify{Cool deal, how do I play this game?}

Battlespace is playable on a local host with multiple players playing against each other or against an agent (Random, Bayesian, Q-learning).
Battlespace can run the game in headless mode with just the agents playing against each other on their own without a server configuration. 
Alternatively, to play a game on localhost, we must create a server using the \texttt{ServerWithUI.py} file. 
After we create a server, we can run two other terminal windows on a local host and connect to that server using the \texttt{HumanInterface.py} file. 
Once the players are connected, they can see a UI grid. 
The system then prompts the players to place their units in their respective playable areas to start the game (Appendix~\ref{appendix:usage} details this process a bit more).

Headless mode is accessible from a separate git branch. 
The users can checkout to `test1' branch and run the scripts starting with \texttt{test\_<agent\_name>.py} for pitting different agents against a Random Agent.

\section{Encoding}
\label{sec:encoding}

\boldify{Now we wrap up and prepare them for the next section on agents}

Based on our initial brainstorming, we started with an encoding approach based on concatenating binary strings.
Over the course of the project, we made changes to this approach and ran experiments with different encodings, eventually arriving at a solution that used many more layers. 
In addition, we also examined an approach used by a built-in agent that uses a list as input.

\subsection{Encoding \#1 - Binary Strings, Few Layers}

\boldify{First we describe the shape of the bucket we are going to fill}

We developed an encoding algorithm to convert observations to 3D tensors of shape $(m, n, numPlayers ^* 2 + 2)$, where $m$ and $n$ are the width and length of the board, respectively. 
Each player has two layers in the third dimension.
Each player's first layer, calculated by $(playerID ^* 2)$, contains the land units of the respective player. 
Each player's second layer, calculated similarly $(playerID ^* 2 + 1)$, contains the air units of the respective player. 
The extra layers encode (1) the obstacles and (2) the actions taken by the players (Note: we did not finish implementing this).
An example of this encoding is available in Appendix~\ref{appendix:ourRepr}.

\boldify{Now we describe the contents of the bucket}

\begin{table}
    \centering
    \begin{tabular}[h] 
    {@{}
    l | 
    p{0.675\linewidth} | 
    p{0.155\linewidth} 
    @{}}
    
    \textbf{Variable} 
    & \textbf{Description} 
    & \textbf{Value Range} 
    \\\hline\hline
    
    \texttt{width} 
    & The width of the board. 
    & \texttt{width} $\in \mathbb{N}$ 
    \\ & \emph{Simplifying Assumption}: $width=10$ (later shrunken further to \texttt{width} $=5$)
    \\\hline
    
    \texttt{length} 
    &  The length of the board.
    & \texttt{length} $\in \mathbb{N}$
    \\ & \emph{Simplifying Assumption}: \texttt{length}$=11$ (later shrunken further to \texttt{length}$=5$)
    \\\hline
    
    \texttt{height} 
    & The height of the board. 
    & \texttt{height} $\in \mathbb{N}$
    \\ & \emph{Simplifying Assumption}: \texttt{height} $=2$ (later shrunken further to \texttt{height} $=1$)
    \\\hline
    
    \texttt{ownerID} 
    & Each team of players receives a team ID starting from 0.
    Walls are not assigned to any team, but projectiles are.
    & \texttt{ownerID} $\in \mathbb{W}$
    \\ & \emph{Simplifying Assumption}: \texttt{ownerID}$\leq 2$
    \\\hline
    
    \texttt{playerID} 
    & Each player receives a player ID starting from 0.
    & \texttt{playerID} $\in \mathbb{W}$
    \\ & \emph{Simplifying Assumption}: \texttt{playerID} $\leq 4$
    \\\hline
    
    \texttt{unitID} 
    & Each unit placed on the board receives an ID, for both movable and immovable objects.
    If new units spawn during the game, for example projectiles, they receive the next counter value.
    Note that under our other simplifying assumptions, \texttt{unitID} is stable between episodes and we can also use it as an indicator of the unit's type.
    & \texttt{unitID} $\in \mathbb{W}$
    \\ & \emph{Simplifying Assumption}: \texttt{unitID} $\leq 32$ (to ensure the binary encoding has a fixed length of 5)
    \\\hline
    
    \texttt{position} 
    &	Position of a unit on the board.
    Each element of the position 3-tuple ranges from \texttt{width}, \texttt{length}, or \texttt{height}, meaning all simplifying assumptions flow downstream from those variables.
    & $(x, y, z),\,\,$where $x, y, z \in \mathbb{W}$
    \\\hline
    
    \texttt{orientation} 
    & Orientation of a unit.
    In Battlespace, most units are only oriented in the plane parallel to the ground.
    The exception is bombs, whose orientation is always $(0, 0, -1)$.
    All other units can take on the eight ordinal orientations: 
    \{(0,1,0), (1,1,0), (-1,1,0), (1,0,0), (-1,0,0), (0,-1,0), (1,-1,0), (-1,-1,0)\}.
    Note that $(0,0,0)$ is not a legal orientation.
    & $(x, y, z), \,\,$where $x,y,z \in \{-1, 0, 1\}$
    \\\hline
    
    \texttt{health} 
    & Represents the health of a unit. 
    & \texttt{health} $\in \mathbb{W}$ 
    \\ & \emph{Simplifying assumption \#1}: \texttt{health} $= 1$ for units which are alive. 
    \\ & \emph{Simplifying assumption \#2}: There are no mechanisms by which to heal, meaning that we need not separately store maximum health and current health.
    \\\hline
    
    \texttt{visibleRange} 
    & Represents the number of squares in each of the eight ordinal directions a unit can see.
    Note that this means diagonal visibility is allowed. 
    & \texttt{visibleRange} $\in \mathbb{W}$ 
    \\ & \emph{Simplifying assumption}: \texttt{visibleRange} $=1$ for all units.
    \\\hline
    \texttt{unitClass} 
    & Binary representation of a unit class (either flag or not flag). If a unit class is of flag type, it is represented by 1 else, it's represented by 0. 
    & \texttt{unitClass} $\in \{0,1\}$ 
    \\\hline
    \end{tabular}
    
    \caption{Table defining variables stored in the game and unit classes, alongside their value ranges.}
    \label{table:variables}

\end{table}

This algorithm takes in the board and calculates a binary string for each cell of the tensor encoding the properties of any unit in that location under the appropriate player's ownership.
To do so, we use the properties found in Table~\ref{table:variables}, such as \texttt{unitId}, \texttt{Orientation}, and \texttt{Health}, following Equation~\ref{eqn:binaryString} to form the binary string and then casting to integer for storage.
\begin{equation}
binaryString = Concat(Binary(unitID), Binary(orientation), Binary(health))
\label{eqn:binaryString}
\end{equation}

\boldify{Next, we describe some simplifying assumptions and the binarization process a little more}

Our implementation makes several simplifying assumptions, all of which are listed in Table~\ref{table:variables}.
In some cases, these assumptions improve the tractability of computation (e.g., board size is relatively small and of fixed size), and in other cases bound the range of numbers to allow the binarization process to generate a fixed length string.
In particular, we capped the \texttt{unitID} to 5 bits, \texttt{orientation} to 4 bits, and \texttt{health} to 1 bit.
This configuration translates to $2^5$ unique possible \texttt{unitID}s, $2^4$ different \texttt{orientation}s, and $2^1$ \texttt{health} values.
The Battlespace domain requires 9 different orientations and we created a dictionary mapping of these orientations to numbers ranging (0--8).

\subsection{Encoding \#2 - Int/Bool, More Layers}
\label{sec:encoding2}

With our encoding strategy, we observed that the \texttt{unitId} was overpowering the \texttt{orientation} while binarizing data.
Upon examining Equation~\ref{eqn:binaryString}, we see that \texttt{unitID} is stored in the most significant bits, while orientation is in the middle bits.
In response, we separated each property into its own layer containing either an integer or a boolean and discarded \texttt{unitID}, meaning each player now has \emph{four} layers in the tensor instead of \emph{two}.
The first layer's two layers represent the player's land units, the first ground layer will contain \texttt{unitID} while the second ground layer will contain \texttt{orientation} in the position of any existing units.
The same approach would work for the air units as well, but instead, we made a simplifying assumption to just use land units and drop air units for this case.

\subsection{Encoding \#3 - list of properties}

Finally, we also experiment with encoding packaged with the Battlespace creators.
An encoded state vector is initialized with properties of the unit like 
['\texttt{owner}', '\texttt{unitID}', '\texttt{unitClass}', '\texttt{visibleRange}', '\texttt{health}', '\texttt{positionX}', '\texttt{positionY}','\texttt{positionZ}', '\texttt{orientationX}', '\texttt{orientationY}', '\texttt{orientationZ}' ].
Then, we check the region of \texttt{visibleRange} for friendly or enemy units.
We encode each square in this range with 0 if no unit is present there, 1 if a friendly unit is present, or 2 if enemy unit is present.
This encoding is appended to the state vector listed previously.
For example, if we have \texttt{visibleRange} of 1, then we have a total of 8 squares around the unit in its visible range to encode.
We repeat the same process is repeated for each playable unit. 

\section{Agents}

\boldify{What is the universe of all agents and who created what}

Battlespace includes a random agent designed to generate random actions.
We then created two additional agents, one based on search and the other based on a NN, which is the focus of this paper.

\subsection{Random Agent} 

\boldify{The random agent has 2 flavors, weighted and uniform, and weighting is useful for various reasons. We will use the random agent inside other things}

The Random Agent is characterized by a list of weights that control the level of preference for each available action. 
Depending on the specified weights, we can have either a uniform random agent or a weighted random agent. 
A Weighted Random Agent might choose actions based on the distribution of human choices (as in \cite{hung2022final}), or based on actions desirable for training because they modify the game state more, like advancing as opposed to doing nothing. 
We used Random Agents when creating other more complex agents, such as when generating rollouts for MCTS (Monte Carlo Tree Search).

\subsection{MCTS Agent}

\boldify{MCTS procedure has the following output and works as follows, at a high level}

We use MCTS (Monte Carlo Tree Search) agents generate a win, loss, and draw probability for each action available from the given state. 
MCTS searches different states in the game by simulating making random moves many times from the current state and recording the results.
Then, once the search tree is constructed, we use it to select an action by taking an arg max based on the recorded win, loss, and draw results.
The core of MCTS is managing the explore-exploit tradeoff to try to focus sampling on parts of the search tree that appear promising.
An alternative that samples more uniformly is called Pure Monte Carlo Game Search (PMCGS)~\cite{russell2016artificial}.

\boldify{We use MCTS 2 ways: the first is an agent. We didn't do very many rollouts for computational reasons, here is an example}

The first component where we used MCTS is the MCTS Agent, which generates rollouts based on (both uniform and weighted) Random Agents' choices and gives the win percentage for each move. 
We experimented with different quantities of rollouts, observing the expected result; as the number of rollouts increases, we see better convergence.
For example, if we have 12 actions and 500 rollouts, each action can be explored 42 times on average, but for the 26 actions for an air unit, that same 500 rollouts only results in around 19 samples for each action. 
Due to computational limitations, we ran experiments for 200 to 1000 rollouts (each game lasts for around 40 to 50 moves on average). 

\boldify{We also use MCTS in the NN agent!, but it is a little different}

The second component where we used MCTS is to provide a training signal for the Neural Network agent, which is responsible for generating the same win, loss, draw probability; but in a different way.
Our goal in this usage is a little different from the MCTS Agent usage, where the goal is to select a good action.
Here, we want the information stored at the top level of the search tree for \emph{all} moves, both good and bad.

\subsection{Neural Network Agent} 

\begin{figure}
    \centering
    \includegraphics[width=\textwidth]{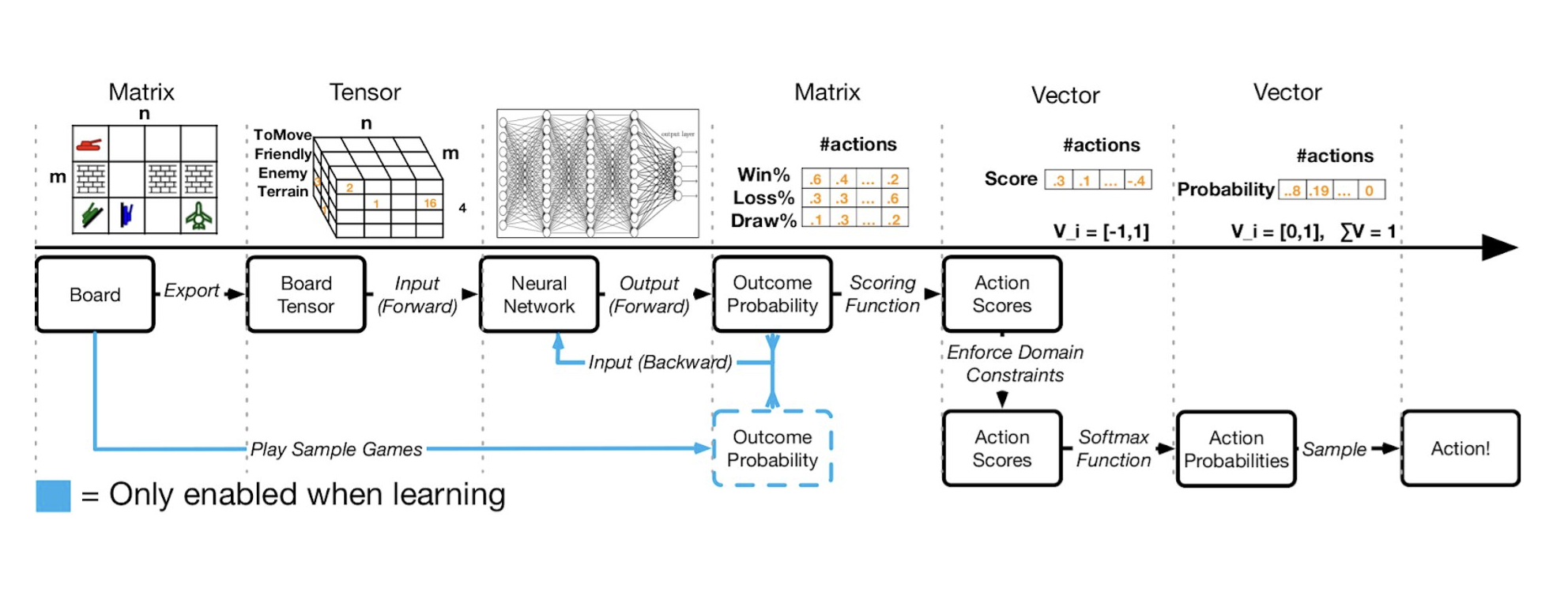}
    \caption{How the AI agent makes a decision, showing nouns in black boxes, verbs on arrows, and the data involved above each step in the pipeline (adapted from \protect\cite{dodge2022mutants}).
	The process begins with a board, which gets converted to a board tensor by separating game objects by faction.
	Here, the green tank is about to move, and we assume the green plane and blue tank are friendly toward the green tank, while the red tank is an enemy, so we encode their type and heading into the tensor.
	A convolutional neural network then takes the board tensor as input and outputs outcome probabilities, which the agent then scores for each action, resulting in a matrix.
	After enforcing domain constraints on the score matrix, the agent applies a softmax to the scores.
	Finally, to select an action, the agent samples from the resulting distribution.
	Parts in {\color{cyan}cyan} activate only during training.
	To contrast with typical approaches, most AI architectures for this task will directly map either \texttt{board} $\rightarrow$ \texttt{action} (an opaque box) or \texttt{(board, action)} $\rightarrow$ \texttt{value} (which must be run many times in order to generate actions/explanations).}
    \label{fig:architecture}
\end{figure}

\boldify{How does the NN agent make a decision? through this series of steps}

Our initial decisionmaking process for the Neural Network Agent is based on the one found in Dodge, et al.~\cite{dodge2022mutants}, as illustrated in Figure~\ref{fig:architecture}.
After running an encoding process on a board, we have a board tensor.
We feed board tensors to the NN, which is trained on the scores calculated by the MCTS procedure. 
We interpret the output from the NN as a set of Outcome Probabilities because that is what the MCTS training signal provides.
In the first step of our post-processing, we put the Outcome Probabilities through a (non-learned) scoring function to calculate Action Scores. 
Next, we enforce domain constraints on the Action Scores by masking illegal moves. 
Last, a softmax function on the Action Scores produces a predicted best move for each playable unit.

\boldify{OK, so you mentioned the guts being this NN, whats the deal there? At a high level, we used DQN, CNN, Dense NN}

Having chosen NN as our function approximator, we experimented with various architectures based on the input encoding.
To train our NNs we rely on the Deep Q-learning Network (DQN) procedure, using a target derived by MCTS. 
Our experiments, detailed in Section~\ref{sec:experiments}, focused primarily on convolutional NN (CNN) and dense NN.
The choice of architecture and hyperparameters can have a substantial impact on the performance of the NN, as it affects the network's ability to capture the relevant features of the board and the game state.

\boldify{Here are some details about tricks we pulled we pulled to stabilize the gradient}

In sequential domains, one common problem when training agents is that the states along a single trajectory are highly correlated.
Thus, in order to reduce unwanted biases, one should seek uncorrelated states, which we used two methods to do.
The first method was to generate a random board, get an action from the agent, perform a simulation step, and a backward pass on the network; then we would throw that board away, generate a new random board, and repeat. 
The second method was the same process with respect to random board generation, but forming mini-batches of size 4--8 before performing the backward pass in an effort to stabilize the gradient. 

\boldify{And how did the our approach shift over time?}

During the course of the project, we made several changes to the architecture, described in Section~\ref{sec:experiments}, based on lessons from experimentation and ideas we had after reading Perolat et al.~\cite{deepNash}.

\section{Experiments}
\label{sec:experiments}

The base version of Battlespace has 4 players competing on a $10\times 11$ grid containing around 30 units (player-deployed units and system-deployed units like walls).
To make our experiments more computationally tractable, we made the simplifying assumptions listed in Table~\ref{table:variables}.
In particular, we reduced the number of units and set the number of players to 2, while also reducing the board size to $5\times5$.
In addition, we sought to avoid wasting computation repeatedly learning from the same (or similar) initial conditions, so we built a random board generator that placed units randomly (both position and orientation) on the board while playing in headless mode. 
This board generator lifted the restriction of units being deployed in the players' own quadrants, meaning that the random board generator can produce states from the mid-game or end-game phase. 

\subsection{Experiments with encoding}

In parallel, we explored multiple encoding techniques among the three described in Section~\ref{sec:encoding} to gain a better understanding of the impact of encoding on the outputs of the neural network.
Encoding \#1 involved converting metadata such as \texttt{unitID}, \texttt{orientation}, and \texttt{health} to binary format, concatenating the bit strings, and then converting the concatenated long bit string to an integer.
We used this integer value as the representation of a particular unit, placing it in the tensor the grid at the unit's position.
However, this method generated some problems due to the concatenation order during the prior conversion.
For instance, since Equation~\ref{eqn:binaryString} illustrates how \texttt{unitID} is contained in the most significant bits, that property received higher importance in the binary-to-integer conversion, while \texttt{health} and \texttt{orientation} received lesser importance.

\subsection{Experiments with neural network}

Our approach was to treat this problem from a computer vision perspective, where we embedded each unit into separate layers and considered these layers as channels of a CNN. 
CNNs are well-suited for the Battlespace domain, because it has a grid structure and most aspects of its rules are translationally-invariant.
CNNs apply small filters (kernels) to local regions of the input data, enabling them to efficiently detect and capture local patterns and features. 
This capability makes CNNs highly beneficial for tasks in domains with partial observability (like Battlespace), where extracting meaningful information from limited visual data is crucial for decision-making and situational awareness.
While experimenting with list encoding, we had to switch to Dense NN because there was no spatial correlation between consecutive elements of an array.

\begin{figure}
\centering
\includegraphics[width=\textwidth]{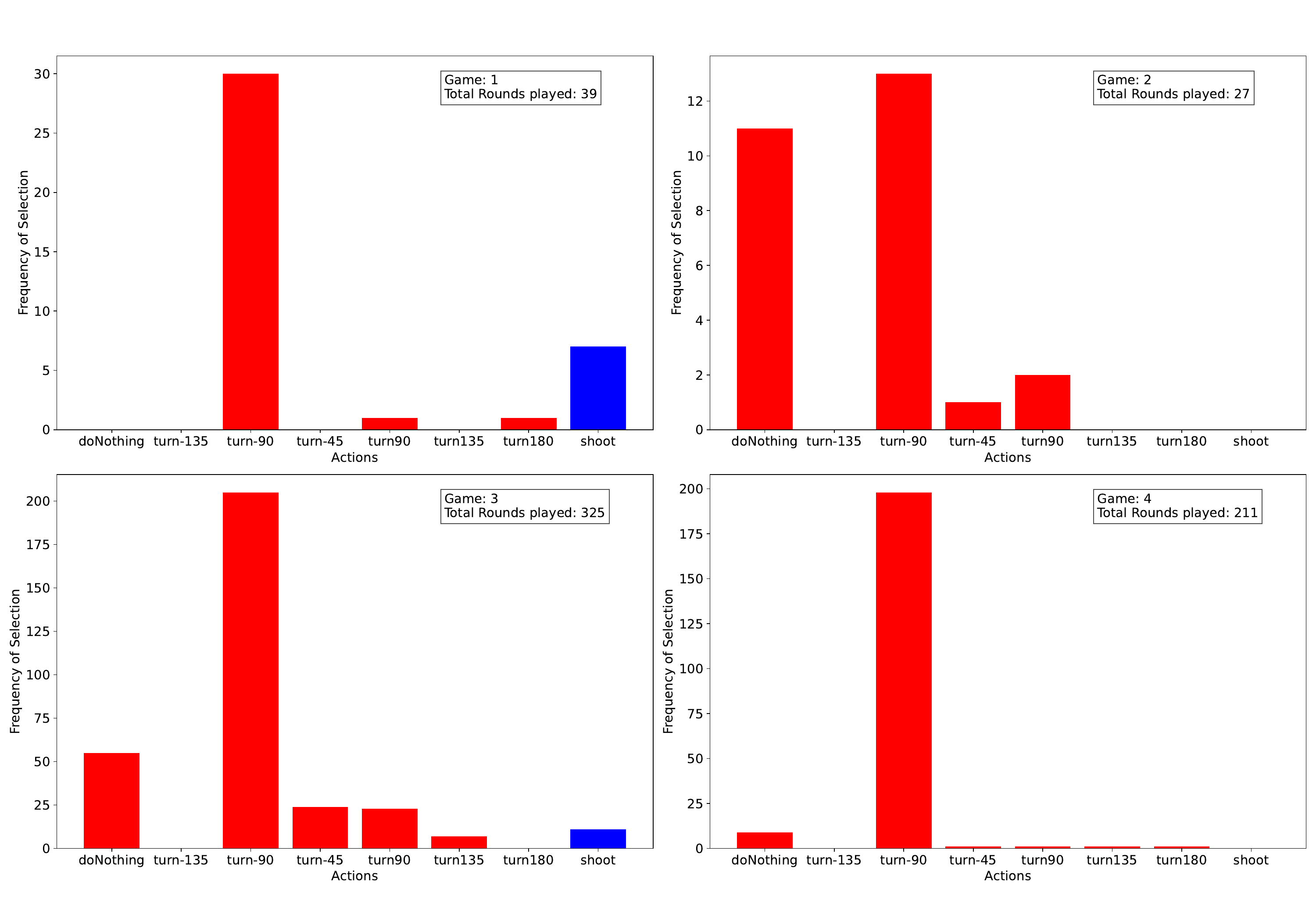}
\caption{Histogram showing actions selected by the NN agent after training for 20 epochs with a batch size of 4 and MCTS rollouts set to 500 in 4 different games.
Each image shows all moves from the start until the game is over, with the number of rounds played in each game shown in the right-hand top corner.
This chart is for the simplified version of the game where the game was played between two players with one land unit per player.
Hence, the number of unique actions is 12.
For better visual clarity some moves (e.g., \texttt{'turn0', 'turn45', 'advance1', 'ram'}) do not appear in the figure because these actions went unused in all of these games.
The actions in red are ``non-impactful'' or non-offensive while the actions in blue are transitional or offensive actions.}
\label{fig:analysis}
\end{figure}

Through our experiments, we observed that the CNN responded poorly to changes in orientation.
To address this issue, we separated the position and orientation into different layers as mentioned in Section~\ref{sec:encoding2}. 
These revisions provide a clearer and more cohesive description of the process and the actions taken during the experimentation.

During our experimentation with different neural network architectures, we made several changes to improve model performance. 
For the CNNs, we explored varying the number of convolutional layers to find the best configuration. 
Additionally, we experimented with adding one or two dense layers after the reshaping operation. 
In the case of Dense Neural Networks, we conducted multiple trials by adjusting the number of layers and layer sizes to determine the optimal depth and width for the architecture.
We used Hare et al.'s~\cite{dod-ai-principles} network as a reference point, drawing insights from its design choices for my experiments with dense networks.
Moreover, the basic CNN architecture from the PyTorch image tutorial served as a starting point, which we customized and modified to suit the specific tasks and datasets at hand. 

\subsection{Experiments with MCTS agent}

The Battlespace domain has state space of $\approx 8\times 10^{514}$, as we detail in Appendix~\ref{appendix:ssc}. 
With our limited computational resources, running MCTS such a large state space proved challenging, even with the speed of generating rollouts quickly with the Random Agent. 
Therefore, we conducted experiments with around 500 rollouts for each board input to the MCTS agent. 
The size of the action space for each state of the game depends on the number of units in play. 
In particular, the total number of unique actions per turn will be a Cartesian product of all the available actions for each unit.
For example, if there are 3 units left on the board and each unit has 12 actions, then the combination of unique actions will be $12^3 = 1728$.
Further simplifying the game, we experimented with just 1 and 2 units per player where the action space sizes were 12 and 144, respectively.

We observed that out of the 12 moves (just considering land units for this example), 9 did not impact the state very much, merely changing the orientation of the unit or doing literally nothing. 
As we were using a Uniform Random Agent during rollouts, low-impact moves were more likely to occur than offensive ones that translated a unit or created a projectile.

The agents learned that sedentary and safe options were better than offensive and risky moves, essentially developing a bias toward safety (Figure~\ref{fig:analysis}).
In this figure, we can see that the action highlighted in red are non-impactful actions like turning and doing nothing.
The ones highlighted in blue are actions that are offensive or change positions like shoot and advance.
We see that most of the time agent prefers to stay in the same position and chooses non-offensive actions.
As a result, it is hard to get the agent into positions that allow it to explore unseen states, and so the agent's grasp of the game dynamics is necessarily limited.

To address the issue with exploration, we experimented with a Non-Uniform Random Agent to generate moves for the MCTS agent, assigning higher weights to actions that create projectiles or adjust positions.
We saw similar results as with Non-Uniform Random Agent.
Specifically, computational challenges to achieve sufficient MCTS rollouts remained our greatest barrier to an agent that selects quality moves.

\section{Discussion - Comparing how each encoding represents the structured data}

In the initial encoding, we latched the variables together by concatenating them into a long string (essentially superimposing the properties).
However, in the subsequent encodings, we placed them one on top of the other (essentially juxtaposing the properties).
In this case, there is a trade-off between complexity in the form of many cells with simple contents vs few cells with complex contents\footnote{%
Consider that most vision applications use either 3 or 4 channel images, while our ``many layers'' encoding uses 6 under our simplifying assumptions, but could require many more than 6 channels.}.
However, we found that employing the first method resulted in agents that prioritized the ordering of the concatenation whereas the second one resulted in agents that prioritized the position of the units in making decisions.
Further, partway through this process we discovered the paper by Perolat et al.~\cite{deepNash}, which used a very similar approach to our Encoding \#2 (detailed later in Appendix Figure~\ref{fig:deepNashEncode}).
Note that both DeepMind's approach and approaches are similar in that the number of layers required grows rapidly as the number and range of unit properties grows, so building more compact representations for structured data remains an open question.
\section{Conclusion and Future Work}

We have described a series of experiments on various types of encodings, architecture, and scoring functions, arriving at  insights for each topic.

Our scoring function relied on calculating win percentages. 
Due to the reduction in unit count many games ended in a draw because of timeouts or mutual destruction.
We did not expect this because draws seem to be rare in human games in the Battlespace domain.

We have tested various convolutional neural networks (CNNs) and observed a substantial difference in outputs before and after resizing the board and re-sampling the number of units.
Our model's losses decrease initially indicating a good training signal.
However, as most of the action space has stationary actions, agents developed a bias for safe plays (i.e., playing to not lose instead of playing to win).
We used outputs from Monte Carlo tree search (MCTS) to generate scores for games.
Our Neural Network learned to predict these scores given a board as input. 
As we increased the number of playable units, the action space grew exponentially, creating a computational bottleneck.
One solution to this bottleneck that we would like to consider in future work is to abandon the search tree entirely and attempt to adapt the game-theoretic approach found in the DeepNash Stratego agent~\cite{deepNash} to this domain.

Our initial encoding approach was to concatenate the binary representation of unit’s metadata information.
This approach demonstrated the importance of concatenation order in the encoding process.
However, we acknowledge the need for further research to improve encoding strategies.
In our current architecture, we have implemented only the state encodings.
We were inspired by the Deep Nash paper and we plan to encode move history in our neural network architecture in future work.

\bibliographystyle{ACM-Reference-Format}
\bibliography{00-paper}


\begin{thebibliography}{8}


\ifx \showCODEN    \undefined \def \showCODEN     #1{\unskip}     \fi
\ifx \showDOI      \undefined \def \showDOI       #1{#1}\fi
\ifx \showISBNx    \undefined \def \showISBNx     #1{\unskip}     \fi
\ifx \showISBNxiii \undefined \def \showISBNxiii  #1{\unskip}     \fi
\ifx \showISSN     \undefined \def \showISSN      #1{\unskip}     \fi
\ifx \showLCCN     \undefined \def \showLCCN      #1{\unskip}     \fi
\ifx \shownote     \undefined \def \shownote      #1{#1}          \fi
\ifx \showarticletitle \undefined \def \showarticletitle #1{#1}   \fi
\ifx \showURL      \undefined \def \showURL       {\relax}        \fi
\providecommand\bibfield[2]{#2}
\providecommand\bibinfo[2]{#2}
\providecommand\natexlab[1]{#1}
\providecommand\showeprint[2][]{arXiv:#2}

\bibitem[Anderson et~al\mbox{.}(2019)]%
        {anderson2019explainingMereMortals}
\bibfield{author}{\bibinfo{person}{Andrew Anderson}, \bibinfo{person}{Jonathan
  Dodge}, \bibinfo{person}{Amrita Sadarangani}, \bibinfo{person}{Zoe
  Juozapaitis}, \bibinfo{person}{Evan Newman}, \bibinfo{person}{Jed Irvine},
  \bibinfo{person}{Souti Chattopadhyay}, \bibinfo{person}{Alan Fern}, {and}
  \bibinfo{person}{Margaret Burnett}.} \bibinfo{year}{2019}\natexlab{}.
\newblock \showarticletitle{Explaining Reinforcement Learning to Mere Mortals:
  An Empirical Study}. In \bibinfo{booktitle}{\emph{International Joint
  Conference on Artificial Intelligence}}. IJCAI, \bibinfo{address}{Macau,
  China}.
\newblock


\bibitem[Anderson et~al\mbox{.}(2020)]%
        {anderson2020mereMortals}
\bibfield{author}{\bibinfo{person}{Andrew Anderson}, \bibinfo{person}{Jonathan
  Dodge}, \bibinfo{person}{Amrita Sadarangani}, \bibinfo{person}{Zoe
  Juozapaitis}, \bibinfo{person}{Evan Newman}, \bibinfo{person}{Jed Irvine},
  \bibinfo{person}{Souti Chattopadhyay}, \bibinfo{person}{Matthew Olson},
  \bibinfo{person}{Alan Fern}, {and} \bibinfo{person}{Margaret Burnett}.}
  \bibinfo{year}{2020}\natexlab{}.
\newblock \showarticletitle{Mental Models of Mere Mortals with Explanations of
  Reinforcement Learning}.
\newblock \bibinfo{journal}{\emph{ACM Trans. Interact. Intell. Syst. (TIIS)}}
  \bibinfo{volume}{10}, \bibinfo{number}{2}, Article \bibinfo{articleno}{15}
  (\bibinfo{year}{2020}), \bibinfo{numpages}{37}~pages.
\newblock
\showISSN{2160-6455}
\urldef\tempurl%
\url{https://doi.org/10.1145/3366485}
\showDOI{\tempurl}


\bibitem[{Defense Innovation Board}(2019)]%
        {dod-ai-principles}
\bibfield{author}{\bibinfo{person}{{Defense Innovation Board}}.}
  \bibinfo{year}{2019}\natexlab{}.
\newblock \showarticletitle{{AI} Principles: Recommendations on the Ethical Use
  of Artificial Intelligence by the {Department of Defense}: Supporting
  Document}.
\newblock \bibinfo{journal}{\emph{United States Department of Defense}}
  (\bibinfo{year}{2019}).
\newblock


\bibitem[Dodge et~al\mbox{.}(2022)]%
        {dodge2022mutants}
\bibfield{author}{\bibinfo{person}{Jonathan Dodge}, \bibinfo{person}{Andrew
  Anderson}, \bibinfo{person}{Matthew Olson}, \bibinfo{person}{Rupika Dikkala},
  {and} \bibinfo{person}{Margaret Burnett}.} \bibinfo{year}{2022}\natexlab{}.
\newblock \showarticletitle{How Do People Rank Multiple Mutant Agents?}. In
  \bibinfo{booktitle}{\emph{ACM Intelligent User Interfaces (IUI)}} (Helsinki,
  Finland). \bibinfo{numpages}{21}~pages.
\newblock
\showISBNx{9781450391443}
\urldef\tempurl%
\url{https://doi.org/10.1145/3490099.3511115}
\showDOI{\tempurl}


\bibitem[Hare et~al\mbox{.}(2021)]%
        {Hare_etal21}
\bibfield{author}{\bibinfo{person}{J~Zachary Hare},
  \bibinfo{person}{B~Christopher Rinderspacher}, \bibinfo{person}{Sue~E Kase},
  \bibinfo{person}{Simon Su}, {and} \bibinfo{person}{Chou~P Hung}.}
  \bibinfo{year}{2021}\natexlab{}.
\newblock \showarticletitle{Battlespace: using {AI} to understand friendly vs.
  hostile decision dynamics in {MDO}}. In \bibinfo{booktitle}{\emph{Artificial
  Intelligence and Machine Learning for Multi-Domain Operations Applications
  III}}, Vol.~\bibinfo{volume}{11746}. International Society for Optics and
  Photonics.
\newblock


\bibitem[Hung et~al\mbox{.}(2022)]%
        {hung2022final}
\bibfield{author}{\bibinfo{person}{Chou~P Hung}, \bibinfo{person}{J~Zachary
  Hare}, \bibinfo{person}{B~Christopher Rinderspacher}, \bibinfo{person}{Sue~E
  Kase}, \bibinfo{person}{Simon~M Su}, \bibinfo{person}{Walter Peregrim},
  \bibinfo{person}{Olena Tkachenko}, \bibinfo{person}{Tomer Krayzman},
  \bibinfo{person}{Adrienne~J Raglin}, {and} \bibinfo{person}{John~T
  Richardson}.} \bibinfo{year}{2022}\natexlab{}.
\newblock \bibinfo{booktitle}{\emph{Final Report of DEVCOM ARL Directors Future
  Ventures (FY21): Novel AI Decision Aids for Decision Dynamics, Deception, and
  Game Theory (Summary Technical Report, Oct 2020-Sep 2021)}}.
\newblock \bibinfo{type}{{T}echnical {R}eport}. \bibinfo{institution}{DEVCOM
  Army Research Laboratory}.
\newblock


\bibitem[Perolat et~al\mbox{.}(2022)]%
        {deepNash}
\bibfield{author}{\bibinfo{person}{Julien Perolat}, \bibinfo{person}{Bart~De
  Vylder}, \bibinfo{person}{Daniel Hennes}, \bibinfo{person}{Eugene Tarassov},
  \bibinfo{person}{Florian Strub}, \bibinfo{person}{Vincent de Boer},
  \bibinfo{person}{Paul Muller}, \bibinfo{person}{Jerome~T. Connor},
  \bibinfo{person}{Neil Burch}, \bibinfo{person}{Thomas Anthony},
  \bibinfo{person}{Stephen McAleer}, \bibinfo{person}{Romuald Elie},
  \bibinfo{person}{Sarah~H. Cen}, \bibinfo{person}{Zhe Wang},
  \bibinfo{person}{Audrunas Gruslys}, \bibinfo{person}{Aleksandra Malysheva},
  \bibinfo{person}{Mina Khan}, \bibinfo{person}{Sherjil Ozair},
  \bibinfo{person}{Finbarr Timbers}, \bibinfo{person}{Toby Pohlen},
  \bibinfo{person}{Tom Eccles}, \bibinfo{person}{Mark Rowland},
  \bibinfo{person}{Marc Lanctot}, \bibinfo{person}{Jean-Baptiste Lespiau},
  \bibinfo{person}{Bilal Piot}, \bibinfo{person}{Shayegan Omidshafiei},
  \bibinfo{person}{Edward Lockhart}, \bibinfo{person}{Laurent Sifre},
  \bibinfo{person}{Nathalie Beauguerlange}, \bibinfo{person}{Remi Munos},
  \bibinfo{person}{David Silver}, \bibinfo{person}{Satinder Singh},
  \bibinfo{person}{Demis Hassabis}, {and} \bibinfo{person}{Karl Tuyls}.}
  \bibinfo{year}{2022}\natexlab{}.
\newblock \showarticletitle{Mastering the game of Stratego with model-free
  multiagent reinforcement learning}.
\newblock \bibinfo{journal}{\emph{Science}} \bibinfo{volume}{378},
  \bibinfo{number}{6623} (\bibinfo{year}{2022}), \bibinfo{pages}{990--996}.
\newblock
\urldef\tempurl%
\url{https://doi.org/10.1126/science.add4679}
\showDOI{\tempurl}
\showeprint{https://www.science.org/doi/pdf/10.1126/science.add4679}


\bibitem[Russell and Norvig(2016)]%
        {russell2016artificial}
\bibfield{author}{\bibinfo{person}{Stuart~J Russell} {and}
  \bibinfo{person}{Peter Norvig}.} \bibinfo{year}{2016}\natexlab{}.
\newblock \bibinfo{booktitle}{\emph{Artificial intelligence: a modern
  approach}}.
\newblock \bibinfo{publisher}{Malaysia; Pearson Education Limited,}.
\newblock


\end{thebibliography}

\newpage
\appendix
\section{State space calculation}
\label{appendix:ssc}

To calculate an estimate of the state space size for the original game including two teams with two players in each team, we begin with the number of objects that might occupy a single square:
\begin{enumerate}
\item 4 soldiers (one of each team's color)
\item 4 tanks
\item 4 trucks
\item 4 flags
\item 4 airplanes
\item $M$ missiles (we estimate that around 6 missiles is reasonable)
\item $B$ bombs (estimate around 1)
\item $W$ walls (estimate around 10)
\item nothing
\end{enumerate}
This results in a total unit count of:
\[totalUnits = 20+M+B+W \approx 37\]

Most of these units can have 8 different orientations, leading to a total possible unit combinations at one square of:
\[combinationsPerSquare = 8^*(20+M)+B+W = 160+8M + B + W \approx 219\]

The board is $10\times11\times2 = 220$ squares, so we have:\\
\[|StateSpace| = combinationsPerSquare^{squares} \approx 219^{220} \approx 8\times 10^{514}\]

At this point, we observe that we discarded \texttt{health}, which would make this estimate would grow even larger.

\clearpage
\section{Game Deployment and Usage}
\label{appendix:usage}

Steps to run the game on server:
\begin{enumerate}
\item Git clone \url{https://gitlab.kitware.com/mixtape/battlespace.git}
\item Install dependencies from the requirement.txt file.
\item Run 3 terminals in parallel and change to the test directory in all of them.
\item Use first terminal to create a server by using the command `\verb+python ServerWithUi.py --test+'.
\item Run the game from second and third terminal by using the command `\verb+python HumanInterface.py+'.
\item Enter the localhost id shown on the first terminal in the second and third terminal to see the deployment phase.
\item Place your units by clicking on the cells in the highlighted grid in yellow (see Appendix Figures~\ref{fig:deployment} and \ref{fig:deployment2}).
\item After all the players place their units, the game begins. Players can enter the action for each of their units base on the prompts shown on their terminal, as shown in Appendix Figure~\ref{fig:humanMove}.
\end{enumerate}

\begin{figure}
\includegraphics[width=\textwidth]{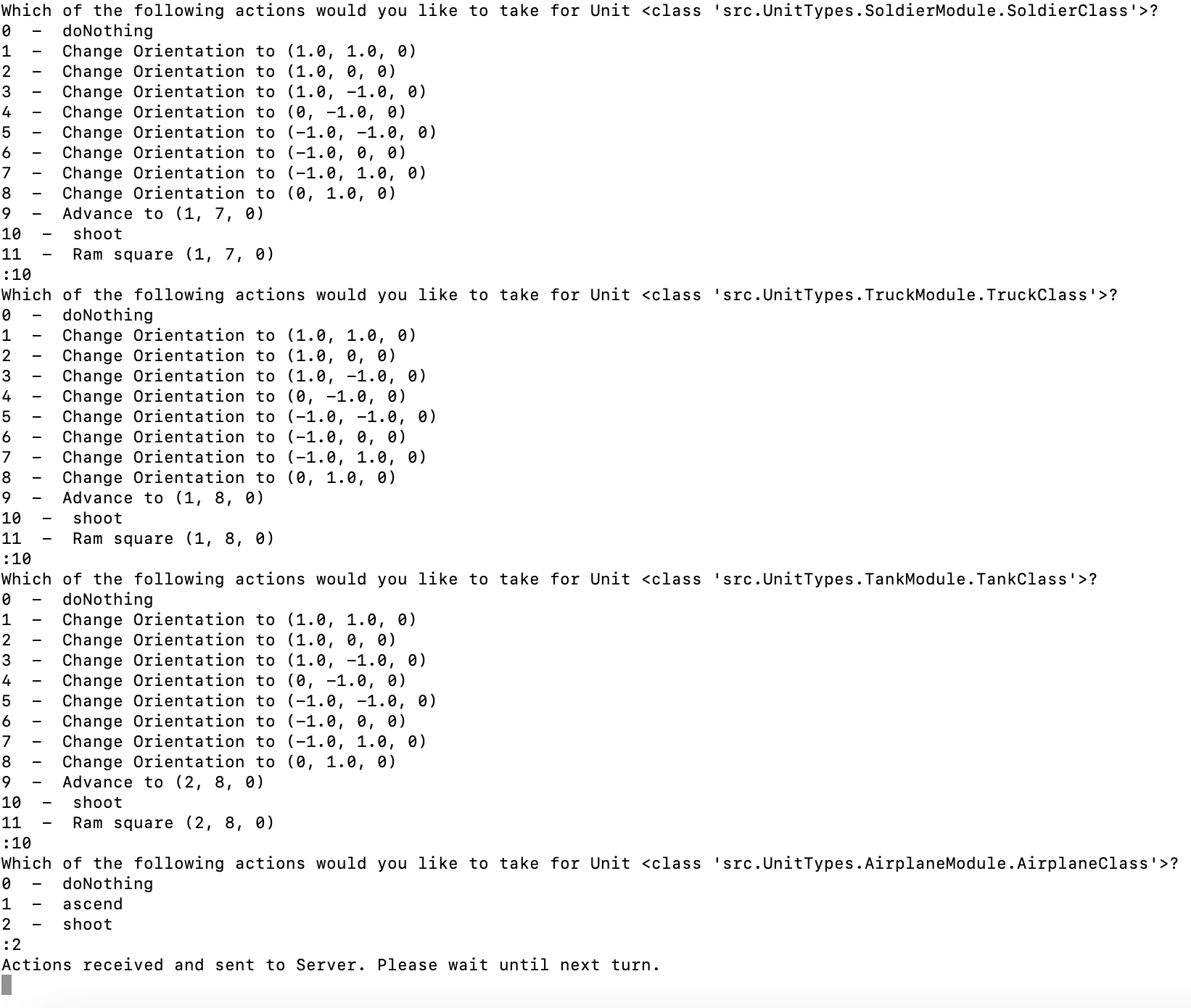}
\caption{A human player makes a move by selecting move options for each playable unit from terminal}
\label{fig:humanMove}
\end{figure}
\begin{figure}
\includegraphics[width=.9\textwidth]{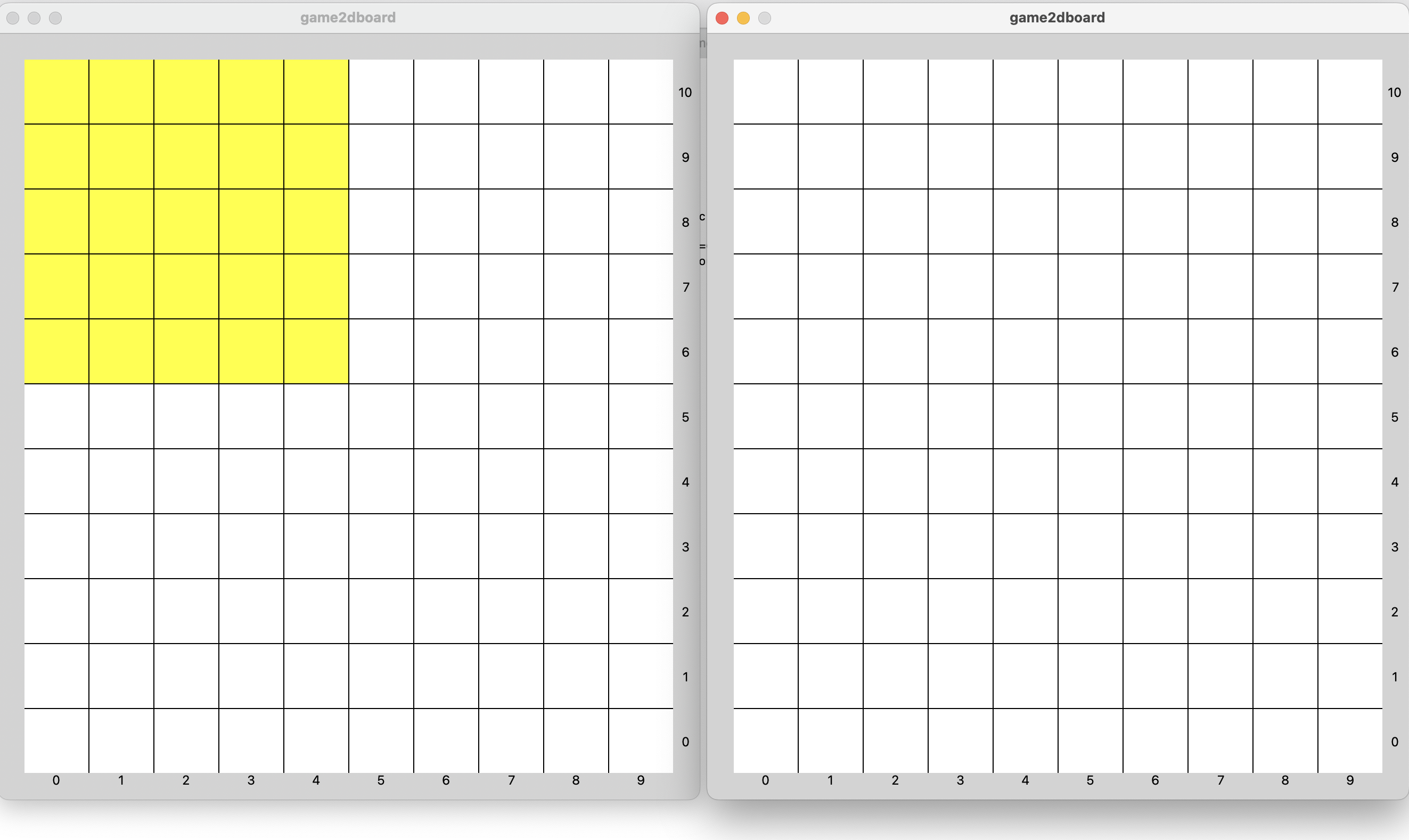}
\caption{Screenshot of Battlespace's UI just prior to deployment phase.
\textbf{Left board}: Player 1 is about to place their units in the region highlighted with yellow.
\textbf{Right board}: Player 2 waits till Player 1 completes deployment.}
\label{fig:deployment}
\end{figure}
\begin{figure}[b]
\includegraphics[width=.9\textwidth]{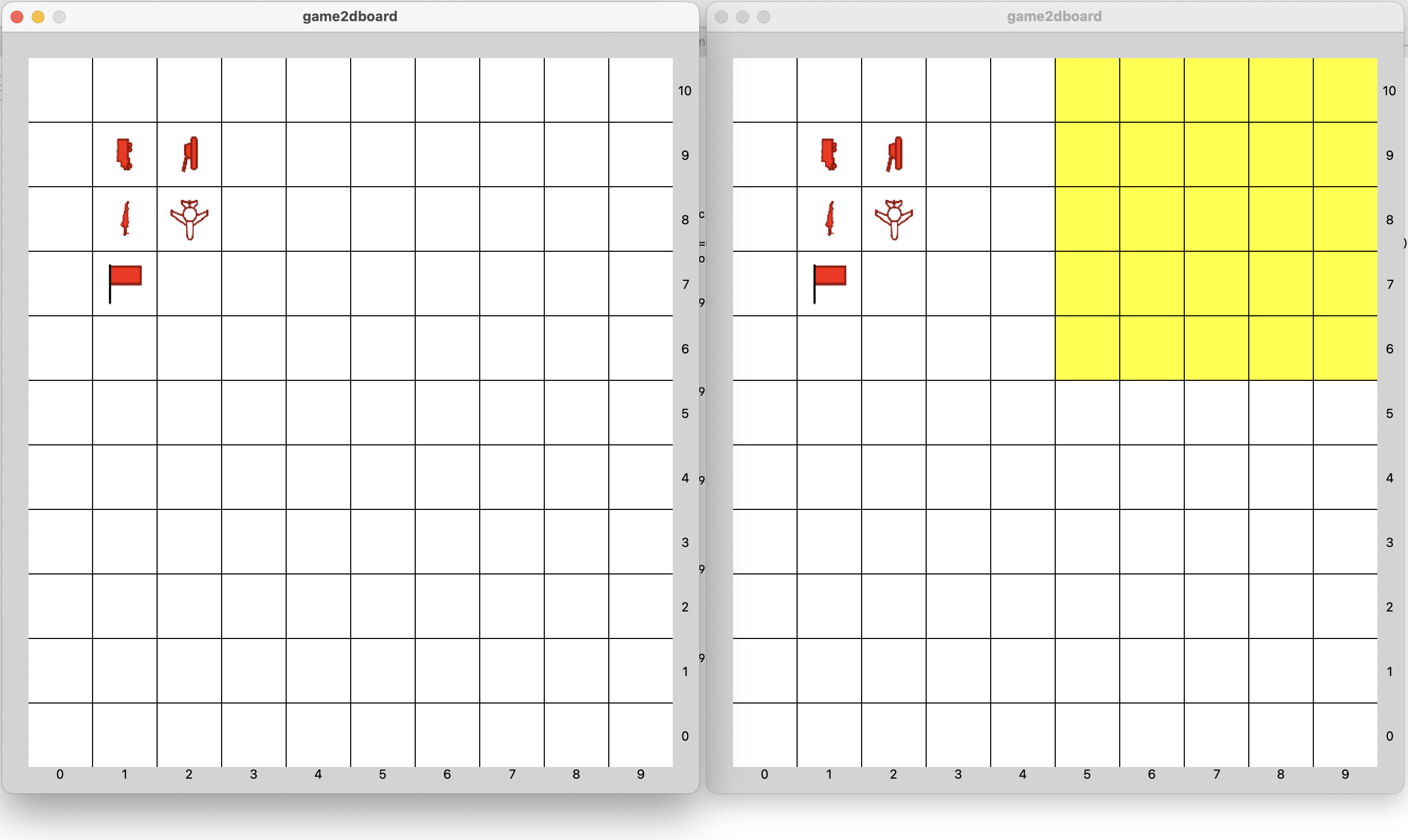}
\caption{Screenshot of Battlespace's UI during the deployment phase.
\textbf{Left board}: Player 1 has placed their units.
\textbf{Right board}: Player 2 initiates deployment.
Note that Player 2 is able to see Player 1's units because they are teammates.
Player 3 will not be able to see either of these choices as they perform deployment.}
\label{fig:deployment2}
\end{figure}
\clearpage
\section{Our Tensor Representation}
\label{appendix:ourRepr}

The following is an example of tensor output from Encoding \#1. 
This example is for the original game running with 10$\times11$ board size, 2 teams, 2 players per team, and 5 units per player. 
The \texttt{tensor\_mapper()} function has embedded the board by using Equation~\ref{eqn:binaryString} to convert the metadata representation of each unit to an integer number which we placed at the same position where the unit is located on the board.
Each player is represented by two layers. 
For example, the first and second layer form the encoding represents the land and air units of Player 1.
The next 6 layers represent the other 3 players. 
The second to last layer represents the walls and the last layer was supposed to encode the previous moves, but we did not finish that aspect of implementation and left it containing zeroes. 
{
\scriptsize
\begin{verbatim}    [[[  0.   0.  75.   0.   0.   0.   0.   0.   0.   0.   0.]
  [  0.   0.   0.   0.   0.   0.   0.   0.   0.   0.   0.]
  [  0. 137.   0.   0.  13.   0.   0.   0.   0.   0.   0.]
  [  0.   0.   0.   0.  37.   0.   0.   0.   0.   0.   0.]
  [  0.   0.   0.   0.   0.   0.   0.   0.   0.   0.   0.]
  [  0.   0.   0.   0.   0.   0.   0.   0.   0.   0.   0.]
  [  0.   0.   0.   0.   0.   0.   0.   0.   0.   0.   0.]
  [  0.   0.   0.   0.   0.   0.   0.   0.   0.   0.   0.]
  [  0.   0.   0.   0.   0.   0.   0.   0.   0.   0.   0.]
  [  0.   0.   0.   0.   0.   0.   0.   0.   0.   0.   0.]]

 [[  0.   0.   0. 111.   0.   0.   0.   0.   0.   0.   0.]
  [  0.   0.   0.   0.   0.   0.   0.   0.   0.   0.   0.]
  [  0.   0.   0.   0.   0.   0.   0.   0.   0.   0.   0.]
  [  0.   0.   0.   0.   0.   0.   0.   0.   0.   0.   0.]
  [  0.   0.   0.   0.   0.   0.   0.   0.   0.   0.   0.]
  [  0.   0.   0.   0.   0.   0.   0.   0.   0.   0.   0.]
  [  0.   0.   0.   0.   0.   0.   0.   0.   0.   0.   0.]
  [  0.   0.   0.   0.   0.   0.   0.   0.   0.   0.   0.]
  [  0.   0.   0.   0.   0.   0.   0.   0.   0.   0.   0.]
  [  0.   0.   0.   0.   0.   0.   0.   0.   0.   0.   0.]]

 [[  0.   0.   0.   0.   0.   0.   0.   0.   0.   0.   0.]
  [  0.   0.   0.   0.   0.   0.   0.   0.   0.   0.   0.]
  [  0.   0.   0.   0.   0.   0.   0.   0.   0.   0.   0.]
  [  0.   0.   0.   0.   0.   0.   0.   0.   0.   0.   0.]
  [  0.   0.   0.   0.   0.   0.   0.   0.   0.   0.   0.]
  [  0.   0.   0. 201. 235.   0.   0.   0.   0.   0.   0.]
  [  0.   0.   0.   0.   0.   0.   0.   0.   0.   0.   0.]
  [  0.   0.   0.   0.   0.   0.   0.   0.   0.   0.   0.]
  [  0. 297.   0.   0.   0.   0.   0.   0.   0.   0.   0.]
  [163.   0.   0.   0.   0.   0.   0.   0.   0.   0.   0.]]

 [[  0.   0.   0.   0.   0.   0.   0.   0.   0.   0.   0.]
  [  0.   0.   0.   0.   0.   0.   0.   0.   0.   0.   0.]
  [  0.   0.   0.   0.   0.   0.   0.   0.   0.   0.   0.]
  [  0.   0.   0.   0.   0.   0.   0.   0.   0.   0.   0.]
  [  0.   0.   0.   0.   0.   0.   0.   0.   0.   0.   0.]
  [  0.   0.   0.   0. 265.   0.   0.   0.   0.   0.   0.]
  [  0.   0.   0.   0.   0.   0.   0.   0.   0.   0.   0.]
  [  0.   0.   0.   0.   0.   0.   0.   0.   0.   0.   0.]
  [  0.   0.   0.   0.   0.   0.   0.   0.   0.   0.   0.]
  [  0.   0.   0.   0.   0.   0.   0.   0.   0.   0.   0.]]

 [[  0.   0.   0.   0.   0.   0.   0.   0.   0.   0.   0.]
  [  0.   0.   0.   0.   0.   0.   0.   0.   0.   0.   0.]
  [  0.   0.   0.   0.   0.   0.   0.   0.   0.   0.   0.]
  [  0.   0.   0.   0.   0.   0.   0.   0.   0.   0.   0.]
  [  0.   0.   0.   0.   0.   0. 457.   0.   0.   0.   0.]
  [  0.   0.   0.   0.   0.   0.   0.   0.   0.   0.   0.]
  [  0.   0.   0.   0.   0.   0.   0.   0.   0.   0.   0.]
  [  0.   0.   0.   0.   0.   0.   0.   0.   0.   0.   0.]
  [  0.   0.   0.   0.   0.   0.   0.   0.   0.   0.   0.]
  [  0.   0.   0.   0.   0.   0.   0.   0.   0.   0.   0.]]

 [[  0.   0.   0.   0.   0.   0.   0.   0.   0.   0.   0.]
  [  0.   0.   0.   0.   0.   0.   0.   0.   0.   0.   0.]
  [  0.   0.   0.   0.   0.   0.   0.   0.   0.   0.   0.]
  [  0.   0.   0.   0.   0.   0.   0.   0.   0.   0.   0.]
  [  0.   0.   0.   0.   0.   0. 423.   0.   0.   0.   0.]
  [  0.   0.   0.   0.   0.   0.   0.   0.   0.   0.   0.]
  [  0.   0.   0.   0.   0.   0.   0.   0.   0.   0.   0.]
  [  0.   0.   0.   0.   0.   0.   0.   0.   0.   0.   0.]
  [  0.   0.   0.   0.   0.   0.   0.   0.   0.   0.   0.]
  [  0.   0.   0.   0.   0.   0.   0.   0.   0.   0.   0.]]

 [[  0.   0.   0.   0.   0.   0.   0.   0.   0.   0.   0.]
  [  0.   0.   0.   0.   0.   0.   0.   0.   0.   0.   0.]
  [  0.   0.   0.   0.   0.   0.   0.   0.   0.   0.   0.]
  [  0.   0.   0.   0.   0.   0.   0.   0.   0.   0.   0.]
  [  0.   0.   0.   0.   0.   0.   0.   0.   0.   0.   0.]
  [  0.   0.   0.   0.   0.   0. 617.   0.   0.   0.   0.]
  [  0.   0.   0.   0.   0.   0.   0.   0.   0.   0.   0.]
  [  0.   0.   0.   0.   0.   0.   0.   0.   0.   0.   0.]
  [  0.   0.   0.   0.   0.   0.   0.   0.   0.   0.   0.]
  [  0.   0.   0.   0.   0.   0.   0.   0.   0.   0.   0.]]

 [[  0.   0.   0.   0.   0.   0.   0.   0.   0.   0.   0.]
  [  0.   0.   0.   0.   0.   0.   0.   0.   0.   0.   0.]
  [  0.   0.   0.   0.   0.   0.   0.   0.   0.   0.   0.]
  [  0.   0.   0.   0.   0.   0.   0.   0.   0.   0.   0.]
  [  0.   0.   0.   0.   0.   0.   0.   0.   0.   0.   0.]
  [  0.   0.   0.   0.   0.   0. 583.   0.   0.   0.   0.]
  [  0.   0.   0.   0.   0.   0.   0.   0.   0.   0.   0.]
  [  0.   0.   0.   0.   0.   0.   0.   0.   0.   0.   0.]
  [  0.   0.   0.   0.   0.   0.   0.   0.   0.   0.   0.]
  [  0.   0.   0.   0.   0.   0.   0.   0.   0.   0.   0.]]

 [[  0.   0.   0.   0.   0. 649.   0.   0.   0.   0.   0.]
  [  0.   0.   0.   0.   0. 681.   0.   0.   0.   0.   0.]
  [  0.   0.   0.   0.   0. 713.   0.   0.   0.   0.   0.]
  [  0.   0.   0.   0.   0. 745.   0.   0.   0.   0.   0.]
  [  0.   0.   0.   0.   0.   0.   0.   0.   0.   0.   0.]
  [  0.   0.   0.   0.   0. 809.   0.   0.   0.   0.   0.]
  [  0.   0.   0.   0.   0. 841.   0.   0.   0.   0.   0.]
  [  0.   0.   0.   0.   0. 873.   0.   0.   0.   0.   0.]
  [  0.   0.   0.   0.   0. 905.   0.   0.   0.   0.   0.]
  [  0.   0.   0.   0.   0. 937.   0.   0.   0.   0.   0.]]

 [[  0.   0.   0.   0.   0.   0.   0.   0.   0.   0.   0.]
  [  0.   0.   0.   0.   0.   0.   0.   0.   0.   0.   0.]
  [  0.   0.   0.   0.   0.   0.   0.   0.   0.   0.   0.]
  [  0.   0.   0.   0.   0.   0.   0.   0.   0.   0.   0.]
  [  0.   0.   0.   0.   0.   0.   0.   0.   0.   0.   0.]
  [  0.   0.   0.   0.   0.   0.   0.   0.   0.   0.   0.]
  [  0.   0.   0.   0.   0.   0.   0.   0.   0.   0.   0.]
  [  0.   0.   0.   0.   0.   0.   0.   0.   0.   0.   0.]
  [  0.   0.   0.   0.   0.   0.   0.   0.   0.   0.   0.]
  [  0.   0.   0.   0.   0.   0.   0.   0.   0.   0.   0.]]]
\end{verbatim}
}

\clearpage
\section{DeepMind's Tensor Representation}
\label{appendix:deepmindTensor}

\begin{figure}
\includegraphics[width=\textwidth]{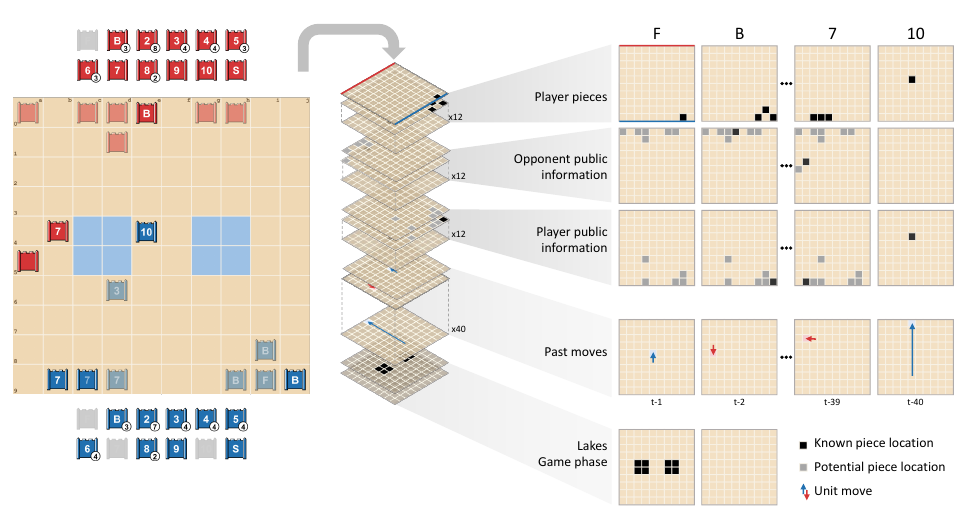}
\caption{\textbf{\emph{(Image and caption credit to Perolat et al.~\protect\cite{deepNash})}}
: The input of the neural network is a single tensor encoding the position of pieces, the currently
known information of both opponent and own pieces (whether a piece moved or was revealed), a limited
move history and the position of the lakes.}
\label{fig:deepNashEncode}
\end{figure}

Figure~\ref{fig:deepNashEncode} illustrates the encoding process DeepMind used in Perolat et al.~\cite{deepNash}.
Note that the approach illustrated here is a mixture of boolean (e.g., player pieces) and integer (e.g., opponent visible information).
Also note that the number of layers required for this approach seems to grow as a Cartesian product of the properties game objects exhibit.

\end{document}